\documentclass[runningheads]{llncs}
\usepackage[T1]{fontenc}
\usepackage{graphicx}
\usepackage{xcolor}
\usepackage{booktabs}
\usepackage{fvextra}
\usepackage{amsmath}
\usepackage{makecell}
\usepackage{multirow}
\usepackage{longtable}
\usepackage{pdflscape} 
\begin{document}
\title{Assessing Readability with LLMs: The Role of Reasoning and Few-Shot Prompting}
\titlerunning{Assessing Readability with LLMs}
%
\author{Raphaël Thieffry\inst{1,2}\orcidID{0009-0008-7988-2945} \and \\
Matej Martinc\inst{2}\orcidID{0000-0002-7384-8112}}

\authorrunning{Raphaël Thieffry and Matej Martinc}

\institute{Université Paris-Saclay, Orsay, France \\
\email{raphael.thieffry@universite-paris-saclay.fr}\\
\and 
Jožef Stefan Institute, Ljubljana, Slovenia \\
\email{matej.martinc@ijs.si}
}


\maketitle 

\begin{abstract}
Readability assessment is essential for tailoring texts to intended audiences across educational, healthcare, and information retrieval domains. However, traditional readability formulas struggle to generalize across genres and languages, while supervised machine learning models rely on scarce, domain-specific annotated corpora, limiting their applicability—particularly for less-resourced languages. Large Language Models (LLMs) offer a highly scalable, multilingual alternative that requires no task-specific training, yet the impact of advanced prompting strategies on their performance remains underexplored. In this paper, we conduct a systematic benchmark of diverse open-source LLMs for multilingual readability assessment, focusing on the prediction of discrete readability levels required by educational frameworks. In addition to English, we evaluate our approach on a less-resourced language, Slovenian, to establish whether LLMs remain effective in low-resource settings. Specifically, we investigate the influence of explicit reasoning, demonstrating that Chain-of-Thought (CoT) prompting and reasoning-oriented models yield significant improvements over direct answering. Furthermore, our exploration of few-shot in-context learning reveals that providing just one labelled example per category (1-shot) substantially enhances prediction quality compared to zero-shot settings, with additional examples offering diminishing returns. By comprehensively comparing these approaches against traditional unsupervised metrics and state-of-the-art supervised baselines, we establish the viability of out-of-the-box LLMs as robust, cross-lingual readability assessors.

\end{abstract}
\section{Introduction}
Readability describes the ease with which a given reader can decode, process and
retain a written text. Because it conditions whether information is effectively
transmitted rather than merely published, measuring it reliably is a practical
concern across a wide range of domains. Generally, readability estimates are used to ensure that resources remain intelligible to their intended audiences \cite{lang2025}. For example, they are used to match educational texts to student and language-learner levels \cite{karaca2024,xia2016}, and to ensure patient comprehension of healthcare materials \cite{bhatt2024,lin2024}. Additionally, readability serves as an evaluation metric for natural language processing systems \cite{picton2025} and as a predictor in various downstream tasks. In information retrieval, readability assessment is used for personalized search, reading-level-aware document re-ranking, and filtering retrieval results to match user expertise or target audience constraints.

The usefulness of these applications rests entirely on the quality of the underlying measure. Readability reflects the cognitive load that a text imposes on a reader, and therefore depends simultaneously on lexical and syntactic sophistication, discourse cohesion, and the background knowledge the reader brings to the text \cite{crossley2017}. Traditional scores transfer poorly across genres \cite{sheehan2013} and languages \cite{madrazo2020}, which limits their validity precisely in the heterogeneous settings where they are most applied. While manual annotation of large corpora provides a more faithful alternative, it is far too slow and costly to be practical at scale \cite{grossman2026}. Consequently, and despite their documented limitations, simplistic readability formulas continue to dominate applied research \cite{grossman2026}.

Recent data-driven approaches address part of this gap but introduce constraints of their own. Supervised models must be trained on annotated corpora that are scarce and usually tied to a specific audience, genre, and labelling scheme, meaning a model trained for one setting rarely transfers to another. Moreover, the overwhelming majority of readability research has been conducted on English, leaving less-resourced languages such as Slovenian with neither the corpora nor the validated tools available for English. To overcome these bottlenecks, recent research has explored measuring readability using large language models (LLMs) \cite{grossman2026}. Because LLMs inherently support multiple languages and require no task-specific training resources, they offer a highly scalable alternative. With specific prompting techniques, LLMs have demonstrated promising results, frequently outperforming traditional readability formulas. Nevertheless, a critical research gap remains: to the best of our knowledge, no prior study has investigated the efficacy of few-shot approaches or the influence of explicit reasoning on LLM-based readability assessment.
 
To address this, we conduct a systematic benchmark of LLMs as readability assessors on multilingual test sets, utilizing high-resource English data and a less-resourced Slovenian dataset to establish cross-lingual generalizability. Rather than proposing a new formula, we ask how well general-purpose models perform out-of-the-box, and under which conditions their predictions become most competitive. We evaluate a range of recent open-source models, including reasoning-oriented ones, and contrast direct answers with chain-of-thought (CoT) prompting. We further examine whether supplying labelled examples for each category—moving from zero-shot to few-shot prompting—improves classification quality, and how quickly this benefit saturates. Finally, we benchmark these prompting strategies against existing baselines to establish where LLMs stand in the current methodological landscape.

Specifically, our main contributions are the following:
\begin{itemize}
\item A multilingual evaluation of several open-source LLMs, varying in size and language coverage, across diverse readability datasets.
\item The first systematic study on the influence of explicit reasoning (via CoT prompting and reasoning-oriented models), showcasing that allowing models to reason before answering yields substantial improvements over direct answering.
\item Exploration of few-shot in-context learning for readability assessment, where we show that a single labelled example per category (1-shot) improves prediction quality substantially over the zero-shot setting, whereas adding further examples brings little to no additional benefit.
\item A comprehensive comparison of these LLM-based approaches against both traditional unsupervised metrics and state-of-the-art (SOTA) supervised methods. Our study is also the first that evaluates LLMs on the task of predicting discrete readability levels, which are usually required by educational frameworks.
\end{itemize}

All code, prompts and evaluation scripts required to reproduce our experiments are publicly available at \url{https://github.com/raphael-thieffry/LLM-Readability-Assessment}.

\section{Related Work}

\paragraph{Formulas and feature-based models.}
Classical readability metrics---FRE, FKGL, ARI, SMOG, Gunning Fog, Spache,
Dale--Chall and LIX---estimate difficulty through linear combinations of surface
proxies, primarily sentence length together with word length or familiar-word
coverage \cite{aiTraditionalMetrics,skvorc2019}. They are transparent and cheap, but insensitive to syntactic depth, discourse cohesion and semantic density
\cite{readabilityReconsidered}. Statistical NLP subsequently recast readability as
supervised prediction over engineered feature vectors spanning lexico-semantic,
syntactic, cohesion and language-model-probability tiers
\cite{martinc2021}.
 
\paragraph{Supervised neural models.}
Transformer encoders learn readability-relevant structure directly from raw text
and outperform feature-based classifiers on standard English and Slovene datasets;
augmenting them with hand-crafted morphological and syntactic features yields no
gain when training data is sufficient, indicating that contextual representations
already encode those properties \cite{martinc2021}. The trade-off is a dependence
on large annotated corpora and a loss of interpretability \cite{martinc2021}.
 
\paragraph{Perplexity and surprisal.}
Probability-based scoring requires no supervision and transfers across languages
and domains \cite{martinc2021,referenceFreeSurvey}, and token-level surprisal is
theoretically grounded in the link between predictability and online processing
effort \cite{hale2001,levy2008,smith2013}. Its validity is nonetheless limited in
two respects: perplexity is anchored in local token prediction and does not track
long-document comprehension \cite{perplexityLongText}, and its relation to
difficulty is non-monotonic, since restricted-vocabulary simplifications diverge
from pretraining distributions and can score higher than advanced academic prose
\cite{martinc2021}.
 
\paragraph{LLMs as readability judges.}
Prompted LLMs are increasingly used to assign grade or CEFR levels without
task-specific training, and are evaluated on corpora such as CLEAR,
OneStopEnglish and CEFR-aligned collections \cite{readabilityReconsidered,trott2024,grossman2026}. Trott and Rivière~\cite{trott2024} first showed that zero-shot prompting of
GPT-4 outperforms traditional formulas, on a single English dataset. That
advantage does not obviously generalise: evaluating the same prompting strategy
across ten open-source models and fourteen datasets, Grossman and Chen~\cite{grossman2026} find it
beats the best formula on only six, with zero-shot LLMs, formulas and
surprisal-based scoring achieving similar average performance. 
 
\paragraph{Slovene.}
English formulas yield invalid absolute grade estimates for Slovene, a
morphologically rich and lower-resourced language. \v{S}kvorc et al.~\cite{skvorc2019} evaluated ten
formulas and eight linguistic criteria on a stratified Slovenian Gigafida corpus with five readability classes, showing that although absolute scores are unreliable, metrics such as LIX and
adapted Dale--Chall separate categories in relative terms. Martinc et al.~\cite{martinc2021}
on the other hand established the neural state of the art for Slovene, testing several supervised neural models on a corpus of Slovenian schoolbooks. 

\paragraph{Gap.}

While recent work has demonstrated the potential of LLMs for readability assessment, existing methodologies remain restricted. For instance, Grossman and Chen~\cite{grossman2026} showed that open-source LLMs can compete with traditional formulas across six languages. However, their approach computes a continuous score based on expected token probabilities, capturing relative readability rather than the \emph{absolute} discrete levels defined by educational curricula. We address this gap by formulating readability assessment as an absolute classification task. We evaluate our models using metrics designed for discrete levels, such as quadratic weighted $\kappa$, and F1-score, while simultaneously measuring correlation to provide a comprehensive comparison with prior work. Furthermore, we break away from the constraints of their narrow experimental setup, which relied on a single zero-shot template and non-reasoning models, by systematically introducing in-context examples (few-shot prompting) and explicit reasoning mechanisms. Finally, we pioneer the application of LLMs to readability measurement in South Slavic languages.

\section{Methodology}
\label{sec:methodology}
 
\subsection{LLM-Based Readability Assessment}
 
We adopt the zero-shot prompting paradigm introduced by Grossman and Chen~\cite{grossman2026}, in
which a general-purpose instruction-tuned LLM is asked to rate a document on the
scale used by the original annotators, and the rating is read off the generated
output. Three properties motivate this choice. First, the procedure requires no
task-specific training, and therefore applies unchanged to target languages and
label schemes for which no annotated training corpus exists. Second, unlike surface formulas, an LLM
conditions on the full document and can in principle register cohesion, discourse
organisation and conceptual density, the dimensions that formulas are known to
ignore \cite{readabilityReconsidered}. Third, the model's weights are unchanged
throughout, so a single model can be applied to datasets differing in language,
genre and granularity without retraining.\footnote{In the few-shot conditions, the prompt
does carry labelled examples, so those conditions are not unsupervised in the
strict sense; what is avoided is gradient-based training, not exposure to labels.}
 
Our work extends this paradigm along a direction the reference study leaves
unexplored. Grossman and Chen~\cite{grossman2026} hold the prompting strategy fixed---a single
zero-shot template, without in-context examples or elicited reasoning---and vary
models and datasets. We instead vary the configuration itself: the surface format of the prompt, the number of in-context examples and the output regime.

\subsection{Prompting Conditions}
\label{sec:prompting_c}
 
Following Grossman and Chen~\cite{grossman2026}, every prompt states the entire rating scale and a definition for each output class, used by the original annotators. This prompt is written in two different formats to assert the results are not biased by these, namely in an XML form and the original prose form. Just for the direct regime (see Section \ref{sec:output_regime}), we additionally test whether omitting just the class definitions or omitting the class definitions and the rating scale in the prompt in the prose format changes the results, as stated in the conclusion of Grossman and Chen~\cite{grossman2026}. This results in altogether four prompts for this regime.\footnote{All used prompts are available at \url{https://github.com/raphael-thieffry/LLM-Readability-Assessment}.} Beyond this, we vary two factors independently.

\paragraph{Number of in-context examples.}
We compare a zero-shot condition, in which the prompt contains only the task
description and the query document, against few-shot conditions in which $k$
labelled examples per class are prepended \cite{brown2020}. We evaluate
$k \in \{0, 1, 2, 3\}$, holding the example pool fixed across models so that
differences are attributable to the model rather than to example selection.
 
\paragraph{Presence of intermediate reasoning.}
We compare direct scoring, in which the model emits a rating without intervening
text, against conditions that elicit an explicit rationale before the rating
\cite{wei2022,kojima2022}. Because contemporary models differ in whether
reasoning is exposed as a distinct output channel, this factor cannot be
realised by a single prompt template; it is instead operationalised through the
four output regimes described below.
 
\subsection{Output Regimes}
\label{sec:output_regime}

A rating has to be recovered from generated text, and free-form generation offers
no guarantee that the text will carry one in a recoverable form. We
avoid the problem structurally: generation is constrained at decoding time
through the structured-output feature of vLLM \cite{kwon2023}, so that a bare
label is drawn from the label set and, where a rationale is wanted, we try different formats described below. A malformed or absent score is therefore
not representable; the only residual failures are exhausting the generation
budget before the predicted class is given, and overflowing the context. We call each
pairing of a prompting strategy with its decoding constraint an \emph{output
regime}, and evaluate four:

\textit{Direct}: the model emits a label with no
        intervening text, the output being constrained to the label set.\footnote{One model complicates the direct regime. \texttt{gpt-oss} always generates
reasoning tokens, and its minimum setting reduces rather than removes them, so it
is allocated a larger generation budget than the other models and is not strictly
in the same condition. We thus don't report the direct regime of \texttt{gpt-oss} since it is not comparable.}

\textit{Elicited reasoning}: for models without a
        native reasoning mode, the model emits a JSON object pairing a free-text
        analysis with a score drawn from the label set, obtaining a rationale
        from models not trained to produce one.
        
\textit{Native reasoning}: for models that emit
        reasoning through a dedicated channel, the rationale and the label are
        separated according to the convention the model was trained to produce.
        
\textit{Two-pass}: the document is analysed in a
        first call, and the analysis supplied alongside the document in a second,
        scoring call, decoupling analysis from scoring at the cost of doubling
        inference.

\subsection{Score Extraction}
 
Grossman and Chen~\cite{grossman2026} compute a readability score as an expected value over the
model's output distribution at the score position, in contrast to the \emph{vanilla score} of Lai et al.~\cite{lai2025}, which takes only the highest-probability token. We use the vanilla score throughout, treating the task as discrete classification over the integer levels defined by the dataset. Constraining the LLM to predict a specific class enables direct comparison against both supervised classifiers trained on the same corpora and traditional readability formulas
(evaluated via Spearman’s rank correlation with the gold labels). While the expected value could in principle also be mapped onto integer levels by rounding, we do not apply this method to preserve comparability across regimes. Under the reasoning regimes the model has already committed to a rating in its rationale before the score position is reached, so the distribution at that position is effectively degenerate and the weighted sum collapses to the arg-max. The expected value is
therefore available only in the direct regime, and adopting it there would confound
the effect of reasoning with a change in how the score is read off. Applying the
vanilla score uniformly keeps the four regimes on a common footing.
 
\section{Experimental Setup}
\label{sec:experimental_setup}
 
\subsection{Models}
 
We evaluate five open-weight instruction-tuned LLMs, listed in
Table~\ref{tab:models}. To select these, we checked for the most popular models published on hugging-face during the years 2025 and 2026 under 32B parameters until June 2026, date of the experiments.
Under these criteria, the final selection allows us to analyse the impact of the different tested variables across a wide range of models ranging from small (Qwen3.5-9B - 9B) to medium sized (GLM-4.7-Flash - 32B).
 
Declared language coverage varies widely across this set. Gemma~3 is pretrained on
over 140 languages, with out-of-the-box support for more than 35
\cite{gemma3}, and Qwen3.5 reports 201 languages and dialects \cite{qwen35}.
At the other end, the \texttt{phi-4} model card states that the model is trained
primarily on English and is not intended for multilingual use, with multilingual
data forming roughly 8\% of the training mixture, and \texttt{gpt-oss} is
likewise documented as trained on a primarily English, text-only corpus. Z.AI
publishes no specific language list for GLM-4.7-Flash, but the model is advertised as multilingual.
 
Grossman and Chen~\cite{grossman2026} restrict their non-English comparisons to models documented
as trained on the target language. We evaluate all five models on all four datasets and report results on Slovenian even for English centric models.
 
\subsection{Datasets}

We evaluate on four readability corpora, summarised in Table~\ref{tab:datasets}:
three English---Newsela \cite{TODOnewsela}, OneStopEnglish \cite{TODOonestop}
and WeeBit \cite{TODOweebit}---and one Slovene corpus of school textbooks,
U\v{c}beniki \cite{martinc2021}. They differ substantially in granularity, from
three ESL proficiency bands to thirteen school grades, and in document length,
whose medians span roughly a factor of five.

\begin{table*}[t]
\centering
\begin{minipage}[t]{0.42\textwidth}
  \centering
  \scriptsize
  \setlength{\tabcolsep}{2.5pt}
  \resizebox{\textwidth}{!}{
  \begin{tabular}{@{}llll@{}}
  \toprule
Model & Params & Reas. & Coverage \\
\midrule
\texttt{gpt-oss-20b}    & 20B MoE & native & En-centric \\
\texttt{phi-4}          & 14B     & none   & En-centric \\
\texttt{gemma-3-27b-it} & 27B     & none   & 140+ langs \\
\texttt{GLM-4.7-Flash}  & 30B MoE & hybrid & Multilingual \\
\texttt{Qwen3.5-9B}     & 9B      & hybrid & 201 langs \\
\bottomrule
  \end{tabular}}
  \caption{Evaluated models. \texttt{GLM-4.7-Flash} has 3B active params. Coverage as declared by developers.}
  \label{tab:models}
\end{minipage}\hfill
\begin{minipage}[t]{0.54\textwidth}
  \centering
  \scriptsize
  \setlength{\tabcolsep}{2.5pt}
  \resizebox{\textwidth}{!}{
  \begin{tabular}{@{}llrrrr@{}}
  \toprule
Dataset & Lg. & Cls. & Texts & Med.\ ch. & Sampled \\
\midrule
Newsela        & en & 11 & 9{,}565  & 4{,}298 & 481 \\
OneStopEnglish & en &  3 &    567   & 4{,}337 & 567 \\
U\v{c}beniki   & sl & 13 & 11{,}930 & 1{,}643 & 559 \\
WeeBit         & en &  5 &  3{,}000 &   873   & 565 \\
\bottomrule
  \end{tabular}}
  \caption{Evaluation datasets. Classes denote US grades 2--12 (Newsela), ESL levels (OneStopEnglish), Slovene grades 1--13 (U\v{c}beniki), and target ages 7--16 (WeeBit).}
  \label{tab:datasets}
\end{minipage}\hfill
\end{table*}

To equalise inference cost across corpora of very different sizes, each dataset is
subsampled to at most 567 documents---the size of the smallest corpus---by
stratified sampling with an equal per-class budget of
$\lfloor 567 / n_{\text{classes}} \rfloor$ under a fixed seed. Newsela is the only
corpus that falls short of the cap: grades 10 and 11 contain just 20 and 2
documents respectively, giving 481 sampled documents. Subsampling is also a
practical necessity: a single pass over 567 documents in all conditions already
requires several days of GPU time.
 
\subsection{Protocol and Evaluation}
 
No parameters are trained at any point: the model is frozen and the only
quantity that changes between conditions is the prompt. For the few-shot conditions we partition each dataset into five folds and draw examples exclusively from the complement of the evaluated fold,
under a fixed seed. A redundant filter additionally excludes the query document by text equality at prompt-build time, which matters where a corpus contains (near) duplicates. 
 
Since a prompt carries $k \times n_{\text{classes}}$ examples, few-shot cost
is markedly uneven across corpora: at $k=3$ a U\v{c}beniki prompt carries 39
example documents against OneStopEnglish's 9. A class containing fewer than $k$ documents contributes all it has.
 
We report weighted $F_1$, quadratic weighted
$\kappa$ and Spearman rank correlation. The ordinal
measure---$\kappa$---distinguishes near misses from
distant ones, which plain $F_1$ cannot; Spearman additionally makes the classical formulas comparable at all: they return an unbounded score rather than a level, and only a rank correlation can
be computed without first mapping that score onto the dataset's scale. Metrics are computed over the predictions that parsed and we report invalid rates separately. Additionally, we pick the best configuration according to the mean Spearman correlation across
the four sampled corpora. This configuration is then run over the complete unsampled corpora, allowing a direct comparison with results for classical formulas and supervised models from the related work \cite{martinc2021}.
 
All experiments were run through a purpose-built harness wrapping vLLM
\cite{kwon2023} and runs were distributed over heterogeneous hardware (2$\times$ NVIDIA A100 or 4$\times$ NVIDIA RTX 4090) Decoding is deterministic in the sense that temperature is fixed at $0$ for all models, prompts and runs.

\section{Results}
\label{sec:results}

\begin{table*}[ht!]
\centering
\caption{Performance comparison of 5 LLMs against 5 readability formulas and 5 readability supervised baselines. Evaluation is conducted on Newsela, OneStopEnglish, Slovenian school books, and WeeBit. Metrics include Spearman correlation ($\rho$), Quadratic Weighted Kappa (QWK), and weighted $F_1$. LLM rows use the XML-tagged prompt format; Rows marked $^{*}$ are measured on stratified samples of each corpus; all others on the complete corpora.}
\label{tab:main_results}
\resizebox{\textwidth}{!}{%
\begin{tabular}{@{} ll c *{12}{c} @{}}
\toprule
\multirow{2}{*}{\textbf{Model}} & \multirow{2}{*}{\textbf{Regime}} & \multirow{2}{*}{\textbf{Shots}} &
\multicolumn{3}{c}{\textbf{Newsela}} &
\multicolumn{3}{c}{\textbf{OneStopEnglish}} &
\multicolumn{3}{c}{\textbf{Učbeniki}} &
\multicolumn{3}{c}{\textbf{WeeBit}} \\
\cmidrule(lr){4-6} \cmidrule(lr){7-9} \cmidrule(lr){10-12} \cmidrule(lr){13-15}
& & & $\rho$ & QWK & F1 & $\rho$ & QWK & F1 & $\rho$ & QWK & F1 & $\rho$ & QWK & F1 \\
\midrule
\multicolumn{15}{l}{\textbf{Baselines: Readability Formulas}} \\
\midrule
Avg.\ sentence length & - & - & .951 & - & - & .597 & - & - & .338 & - & - & .649 & - & - \\
FKGL & - & - & .908 & - & - & .561 & - & - & .353 & - & - & .612 & - & - \\
ARI & - & - & .920 & - & - & .555 & - & - & .349 & - & - & .558 & - & - \\
SMOG & - & - & .867 & - & - & .526 & - & - & .437 & - & - & .562 & - & - \\
Dale--Chall & - & - & .773 & - & - & .493 & - & - & .395 & - & - & .442 & - & - \\
\midrule
\multicolumn{15}{l}{\textbf{Baselines: Supervised Models}} \\
\midrule
BERT \cite{martinc2021} & - & - & - & .979 & .751 & - & .708 & .677 & - & .885 & .416 & - & .953 & .858 \\
HAN \cite{martinc2021} & - & - & - & .984 & .810 & - & .825 & .789 & - & .807 & .485 & - & .886 & .752 \\
BiLSTM \cite{martinc2021} & - & - & - & .963 & .699 & - & .723 & .692 & - & .798 & .522 & - & .906 & .775 \\
\midrule
\midrule
\multicolumn{15}{l}{\textbf{Large Language Models}} \\
\midrule
\multirow{2}{*}{gpt-oss-20b} 
& Native Reasoning   & 0$^{*}$ & .791 & .318 & .036 & .356 & .132 & .211 & .696 & .540 & .115 & .520 & .314 & .221 \\
& Two-pass           & 0$^{*}$ & .705 & .268 & .035 & .145 & .040 & .205 & .638 & .461 & .080 & .424 & .247 & .201 \\
\midrule
\multirow{6}{*}{phi-4} 
& \multirow{4}{*}{Direct}    
  & 0$^{*}$ & .841 & .479 & .061 & .131 & .017 & .168 & .699 & .588 & .067 & .328 & .113 & .124 \\
& & 1$^{*}$ & .666 & .400 & .048 & .369 & .183 & .270 & .696 & .587 & .083 & .498 & .356 & .215 \\
& & 2$^{*}$ & .613 & .423 & .095 & .290 & .124 & .249 & - & - & - & .653 & .571 & .313 \\
& & 3$^{*}$ & .608 & .489 & .101 & .141 & .036 & .200 & - & - & - & .666 & .621 & .393 \\
\cmidrule{2-15}
& Elicited Reasoning & 0$^{*}$ & .787 & .371 & .044 & .117 & .011 & .170 & .615 & .436 & .106 & .610 & .412 & .242 \\
& Two-pass           & 0$^{*}$ & .791 & .415 & .037 & - & .000 & .167 & .749 & .410 & .079 & .537 & .353 & .247 \\
\midrule
\multirow{6}{*}{gemma-3-27b-it} 
& \multirow{4}{*}{Direct}    
  & 0$^{*}$ & .810 & .426 & .014 & .147 & .017 & .168 & .819 & .639 & .079 & .581 & .332 & .178 \\
& & 1$^{*}$ & .804 & .590 & .075 & .572 & .412 & .351 & .861 & .785 & .152 & .677 & .632 & .360 \\
& & 2$^{*}$ & .791 & .586 & .095 & .565 & .404 & .351 & \textbf{.884} & .849 & .194 & .719 & .699 & .466 \\
& & 3$^{*}$ & .787 & .544 & .066 & .567 & .368 & .349 & .856 & .799 & .190 & .767 & \textbf{.754} & \textbf{.543} \\
\cmidrule{2-15}
& Elicited Reasoning & 0$^{*}$ & .807 & .411 & .019 & .483 & .192 & .205 & .675 & .469 & .091 & .580 & .376 & .170 \\
& Two-pass           & 0$^{*}$ & .817 & .384 & .017 & .523 & .227 & .213 & .598 & .337 & .058 & .584 & .354 & .154 \\
\midrule
\multirow{6}{*}{GLM-4.7-Flash} 
& \multirow{4}{*}{Direct}    
  & 0$^{*}$ & .786 & .617 & .103 & .440 & .337 & .348 & .617 & .574 & .093 & .495 & .376 & .151 \\
& & 1$^{*}$ & .685 & .441 & .151 & .330 & .226 & .330 & .722 & .645 & .183 & .621 & .538 & .285 \\
& & 2$^{*}$ & .716 & .440 & .157 & .313 & .239 & .345 & .752 & .661 & .213 & .711 & .639 & .367 \\
& & 3$^{*}$ & .697 & .441 & .131 & .344 & .277 & .350 & .764 & .660 & .224 & .731 & .681 & .388 \\
\cmidrule{2-15}
& Native Reasoning   & 0$^{*}$ & .849 & .718 & .174 & .527 & .403 & .370 & .811 & .812 & .191 & .737 & .639 & .345 \\
& Two-pass           & 0$^{*}$ & .859 & .773 & .196 & .537 & .403 & .369 & .698 & .707 & .161 & .625 & .539 & .237 \\
\midrule
\multirow{9}{*}{Qwen3.5-9B} 
& \multirow{4}{*}{Direct}    
  & 0$^{*}$ & .679 & .341 & .033 & .359 & .129 & .212 & .755 & .511 & .057 & .465 & .222 & .119 \\
& & 1$^{*}$ & .704 & .531 & .118 & .528 & .367 & .337 & .772 & .724 & .171 & .644 & .455 & .240 \\
& & 2$^{*}$ & .705 & .651 & \textbf{.198} & .476 & .370 & .396 & .794 & .753 & .217 & .719 & .653 & .438 \\
& & 3$^{*}$ & .681 & .675 & .185 & .387 & .334 & \textbf{.417} & .808 & .790 & .260 & \textbf{.769} & .739 & .483 \\
\cmidrule{2-15}
& Native Reasoning   & 0$^{*}$ & \textbf{.889} & .591 & .043 & \textbf{.640} & .440 & .351 & .876 & .863 & .240 & .680 & .549 & .268 \\
& Two-pass           & 0$^{*}$ & .745 & .536 & .087 & .383 & .289 & .334 & .681 & .658 & .159 & .568 & .369 & .191 \\\hline
Qwen3.5-9B & \textbf{Native reasoning (entire corpora)}   & 1 & .788 & .698 & .148 & .514 & .393 & .357 & .753 & .741 & .252 & .719 & .656 & .322 \\
\bottomrule
\end{tabular}%
}
\end{table*}

The results of our experiments are presented in Table~\ref{tab:main_results}. We only present results for the XML-tagged prompt templates with class definitions to preserve space. Comparing them to the results of the prose templates of
Grossman and Chen~\cite{grossman2026}, model by model and corpus by corpus at every $k$ across all
four output regimes, the two are indistinguishable in rank agreement: the prose template scores $0.008$ lower in $\rho$ on average. The XML-tagged prompt templates nevertheless hold a slight edge in $\kappa$ ($+0.026$) and weighted $F_1$ ($+0.015$). 

For the direct regime, we additionally test two formats : the terse prompt template and the Grossman and Chen~\cite{grossman2026} prompt without level definitions. The terse template, stripped of both structure and level definitions, trails all three on every metric, losing $0.107$ in $\rho$ and $0.113$ in $\kappa$ against
the XML template. Removing the definitions alone --- the two Grossman and Chen~\cite{grossman2026} templates against
each other, the only contrast in which nothing else changes --- costs every model
and every corpus on both ordinal measures. Mean $\rho$ falls from $0.598$ to $0.465$ for
\texttt{qwen3.5-9b}, from $0.556$ to $0.490$ for \texttt{glm-4.7-flash}, from
$0.727$ to $0.670$ for \texttt{gemma-3-27b-it} and from $0.537$ to $0.524$ for
\texttt{phi-4}; in $\kappa$ the losses run from $0.036$ to $0.116$, and per
corpus from $0.032$ on Newsela to $0.110$ on WeeBit. The ablation of Grossman and Chen~\cite{grossman2026} therefore reproduces on a different model set and, for the
first time, on Slovene.

\subsection{Output regime}
\label{sec:regimes}

Reasoning helps, but only when the model produces it natively. Taking the
native-reasoning configurations and comparing each against the direct
configuration of the same model, corpus and number of examples, agreement
improves on every corpus and every measure: averaged over the four corpora,
$+0.136$ in $\rho$, $+0.186$ in $\kappa$ and $+0.086$ in weighted $F_1$
(Table~\ref{tab:regime_gains}). On U\v{c}beniki the two models rise from
$\rho = 0.716$ to $0.855$ and from $\kappa = 0.614$ to $0.850$.

Neither regime that extracts a rationale from a model without a reasoning
channel reproduces this. Elicited reasoning gains $0.006$ in $\rho$ and loses
$0.027$ in $\kappa$; the two-pass regime, at twice the inference cost, loses on
$\kappa$ and weighted $F_1$ on three corpora out of four. Both are at their worst
on U\v{c}beniki, the corpus with the most classes. Obtaining a rationale by prompt is not
a substitute for a model trained to produce one.

The advantage also narrows as examples are added: the native-reasoning gain in
$\rho$ falls from $+0.176$ at $k=0$ to $+0.116$ at $k=3$, and the two-pass regime
turns from $+0.058$ to $-0.037$ over the same range --- a first indication of the
substitution examined in Section~\ref{sec:shots}.\footnote{\texttt{gpt-oss-20b} is absent
throughout, as it is incompatible with the direct regime as explained in Section~\ref{sec:output_regime}).}

\begin{table*}[t]
\centering
\small
\setlength{\tabcolsep}{4pt}
\resizebox{\textwidth}{!}{
\begin{tabular}{@{}lrrrrrrrrrrrrrrr@{}}
\toprule
\multirow{2}{*}{Regime} & \multicolumn{3}{c}{Newsela} & \multicolumn{3}{c}{OneStopEnglish} & \multicolumn{3}{c}{Učbeniki} & \multicolumn{3}{c}{WeeBit} & \multicolumn{3}{c}{Mean} \\
\cmidrule(lr){2-4} \cmidrule(lr){5-7} \cmidrule(lr){8-10} \cmidrule(lr){11-13} \cmidrule(lr){14-16}
& $\Delta\rho$ & $\Delta\kappa$ & $\Delta F_1$ & $\Delta\rho$ & $\Delta\kappa$ & $\Delta F_1$ & $\Delta\rho$ & $\Delta\kappa$ & $\Delta F_1$ & $\Delta\rho$ & $\Delta\kappa$ & $\Delta F_1$ & $\Delta\rho$ & $\Delta\kappa$ & $\Delta F_1$ \\
\midrule
Native reas. & $+.140$ & $+.239$ & $+.084$ & $+.177$ & $+.147$ & $+.038$ & $+.138$ & $+.237$ & $+.126$ & $+.088$ & $+.119$ & $+.096$ & $+.136$ & $+.186$ & $+.086$ \\
Elicited reas. & $+.026$ & $-.023$ & $-.022$ & $+.028$ & $-.046$ & $-.050$ & $-.049$ & $-.070$ & $+.028$ & $+.019$ & $+.033$ & $+.007$ & $+.006$ & $-.027$ & $-.009$ \\
Two-pass & $+.075$ & $+.041$ & $-.015$ & $+.030$ & $-.001$ & $-.026$ & $-.072$ & $-.140$ & $-.057$ & $-.047$ & $-.081$ & $-.068$ & $-.004$ & $-.045$ & $-.042$ \\
\bottomrule
\end{tabular}}
\caption{Change against the direct regime, averaged over models and over
$k \in \{0,1,2,3\}$. Each regime is compared against the direct condition of the
same models, which differ between regimes: native reasoning covers the two models
exposing a reasoning channel and elicited reasoning the two that do not, while
the two-pass regime covers all four.}
\label{tab:regime_gains}
\end{table*}

\subsection{Number of in-context examples}
\label{sec:shots}

The first example per class is worth more than any that follow. Averaged over
models, regimes and the four corpora, $\rho$ rises from $0.603$ at $k=0$ to
$0.646$ at $k=1$ and then flattens ($0.640$, $0.634$), and $\kappa$ behaves the
same way. That average conceals two opposite trends: on Newsela, whose documents are the longest, $\rho$ falls at every step from $0.797$ to $0.751$, while on WeeBit, whose documents are the shortest, it
rises at every step from $0.553$ to $0.642$; OneStopEnglish and U\v{c}beniki both
peak at $k=1$. Exact accuracy, by contrast, keeps rising on every corpus over the
whole range, as does weighted $F_1$. Further examples are therefore not inert:
they sharpen the choice of an individual level while ceasing to improve the ordering.

How much that first example is worth depends on how much reasoning the model is
already doing. Measuring the change from $k=0$ to $k=1$ within each model, corpus
and regime gives $+0.080$ in $\rho$ for direct scoring and $+0.063$ for elicited
reasoning, against $+0.016$ for native reasoning and $+0.015$ for the two-pass
regime. The ordering is the same on every metric, though its magnitude is not:
the gap between direct scoring and native reasoning is a factor of five in
$\rho$ but around $1.3$ in $\kappa$, weighted $F_1$ and exact accuracy. An
example therefore substitutes for reasoning mainly in ranking documents, not in
placing them on the scale --- which is why the highest-ranked configuration and
the next differ only in whether an example is present.

\subsection{Slovene}
\label{sec:slovene}

Two of the five models are not presented by their developers as multilingual, so
Slovene is reported separately, at $k=0$, the only setting in which every model
and regime clears the validity rule on this corpus. Native reasoning dominates:
\texttt{qwen3.5-9b} reaches $\rho = 0.876$ and \texttt{glm-4.7-flash} $0.811$,
against $0.755$ and $0.617$ for the same models scoring directly. The two models
without a reasoning channel are not helped by an elicited rationale here --- it is
their weakest regime, \texttt{gemma-3-27b-it} falling from $0.819$ to $0.675$ and
\texttt{phi-4} from $0.699$ to $0.615$ --- and \texttt{gemma} scoring directly
reaches $0.819$, second only to \texttt{qwen}'s native reasoning. Declared
multilingual coverage predicts the ranking closely: the three models documented as
multilingual lead both English-centric ones on rank agreement and on $\kappa$,
and \texttt{gpt-oss}, the strongest of the two at $\rho = 0.696$, stays below the
direct condition of \texttt{gemma} and \texttt{qwen}.

\subsection{The selected configuration}
\label{sec:selected}

According to the mean $\rho$, the best configuration is \texttt{qwen3.5-9b} with native reasoning and one example
per class. Run on the complete corpora, the selected configuration reaches $\rho = 0.788$ on
Newsela, $0.514$ on OneStopEnglish, $0.753$ on U\v{c}beniki and $0.719$ on
WeeBit, for a mean of $0.694$ against $0.772$ on the samples. Note that these results are not directly comparable to those on the sampled dataset due to the stratified nature of the samples, where classes are balanced. We can nevertheless compare these results directly to the results for readability formulas, which are computed over the complete corpora. Formula scores are oriented so that a higher value always denotes a harder text. The strongest
formula reaches $\rho = 0.948$ on Newsela, $0.595$ on OneStopEnglish, $0.442$ on
U\v{c}beniki and $0.649$ on WeeBit, against $0.788$, $0.514$, $0.753$ and $0.719$ for
the selected configuration. Average sentence length is the strongest formula on
the three English corpora; on Slovene it is the weakest of the five and SMOG
leads instead. The selected configuration therefore trails the best formula on
both parallel corpora and leads on the two others, by $0.070$ on WeeBit and by
$0.311$ on U\v{c}beniki, where every formula is weak.

Newsela deserves a caveat wherever a formula is competitive on it. The corpus is
built by rewriting a single article at several grade levels, and shortening
sentences is among the principal operations that rewriting performs, so average
sentence length approximates a readout of the quantity the annotation process
manipulated. OneStopEnglish is built by the same procedure, which makes it the
control: any gap between the two is specific to how Newsela was rewritten rather
than general to parallel corpora. The gap is large --- average sentence length
reaches $0.948$ on Newsela against $0.595$ on OneStopEnglish.

We also report supervised results on the corpora from Martinc et al.~\cite{martinc2021}. Those models are trained on
part of each corpus while ours sees none of it, so the comparison is between a
supervised ceiling and an untrained method rather than between equals. The best
supervised model reaches $\kappa = 0.885$ against $0.741$ for the selected
configuration on U\v{c}beniki, and $0.953$ against $0.656$ on WeeBit. The gap is
wider on weighted $F_1$ --- $0.522$ against $0.252$ and $0.858$ against $0.322$ ---
since a supervised classifier learns the label distribution of the corpus it was
trained on, which no prompted model has access to.

\subsection{Validity and cost}
\label{sec:validity}

Twelve combinations exceed the 50\% fail threshold and are reported as failures (e.g., as empty cells in Table \ref{tab:main_results}). All twelve are on U\v{c}beniki, and all
occur at $k \geq 1$. An invalid prediction means the generation budget was exhausted or the context overflowed.
The overall invalid rate is approximately 1\%, rising monotonically with the gold
level, from 0.4\% at the easiest to 1.4\% at the hardest: the excluded documents
are mildly biased towards difficult ones. Direct scoring is at zero on four of
the five models; the reasoning arms carry the rest. 

\section{Conclusion}

We benchmarked five open-weight LLMs as readability assessors across four corpora
in two languages, crossing four output regimes with the prompt format and the
number of in-context examples per class, and evaluated the resulting 160
configurations as classifiers over the discrete levels each corpus defines rather
than as producers of a continuous score.

Two findings hold across corpora and metrics. First, reasoning helps only when
the model produces it natively. Second, the first labelled
example per class carries most of the benefit of in-context learning, and what it
contributes depends on what the model is already doing: it improves rank
agreement five times more for a model scoring directly than for one already
reasoning, while further examples continue to sharpen the assignment of absolute
levels without improving the ordering. 

Against external baselines the picture is deliberately mixed. The selected
configuration --- \texttt{qwen3.5-9b} with native reasoning and one example ---
outperforms every classical formula on the Slovene textbook corpus and on WeeBit, but trails average sentence length on the two corpora built by rewriting a single article across grade levels. It
remains below supervised encoders trained on these corpora. The practical reading is
therefore narrow but concrete: where a target language or label scheme has no
annotated corpus --- the situation that motivates the approach --- a small
open-weight reasoning model with a single example per class is the strongest
training-free option we measured, and the gains concentrate exactly where formulas
are weakest. Extending this to further low-resource languages, and quantifying
the inference cost such an assessor imposes when applied at retrieval scale,
are the natural next steps.

\section{Limitations}
\label{sec:limitations}

\begin{enumerate}
    \item \textbf{Stability of the reasoning regimes across runs.} Temperature is
    $0$ throughout and no sampling is performed, but this does not make a long
    rationale stable across serving configurations. OneStopEnglish is the only
    corpus whose sample is the entire corpus, so the selected configuration is
    scored there on the same 567 documents in both the ranking run and the
    full-corpus run, which used different context lengths and hardware. The two
    disagree on 11.3\% of the zero-shot predictions, moving $\rho$ by $0.049$,
    and on 23.1\% at $k=1$, moving $\rho$ by $0.122$. We attribute this to
    numerical differences propagating through hundreds of generated tokens, though
    we cannot exclude an unrecorded difference in the prompt. Either way,
    differences between configurations smaller than this should not be read as
    effects, which is why we treat the two top-ranked configurations as tied.

    \item \textbf{Context truncation and invalid predictions.} A prompt carries
    $k \times n_{\text{classes}}$ examples, so few-shot prompts press against the
    context budget on corpora that combine long documents with many classes: at
    least one prompt truncated the query document for 25.2\% of Newsela and
    26.3\% of U\v{c}beniki documents, against 7.4\% on OneStopEnglish and none on
    WeeBit, whose documents are the shortest. All twelve configuration--corpus
    cells excluded by the 50\% validity rule are Slovene at $k \geq 1$. The
    full-corpus run is affected too: 19.0\% of Newsela and 9.3\% of U\v{c}beniki
    predictions are invalid, against roughly 1\% across the sampled grid, and
    invalid predictions are mildly biased towards harder documents. Full-corpus
    figures are therefore computed over the documents that were answered, and the
    cause lies in the harness rather than in the models.

    \item \textbf{Transfer from samples to complete corpora.} Configurations are
    ranked on stratified samples of at most 567 documents and only the winner is
    run in full, so the ranking itself is never verified at scale. Mean $\rho$
    falls from $0.772$ on the samples to $0.694$ on the complete corpora, and we
    cannot rule out that a configuration ranked lower on the samples would
    transfer better.

    \item \textbf{Evaluation target.} We measure agreement with human-assigned
    pedagogical levels, not reader processing effort. Work using eye-tracking
    indices finds that formulas, commercial systems and LLM judges alike predict
    online reading effort poorly \cite{readingEasePoorPredictors}; our results
    speak to the first target and not the second.
\end{enumerate}

\section{Acknowledgments}

The authors acknowledge the financial support from the Slovenian Research and Innovation Agency – research project No. J5-50169: Linguistic Accessibility of Social Assistance Rights in Slovenia, and core programme funding No. P2-0103.

%
%
%

\bibliographystyle{splncs04}
\bibliography{bibliography}

\end{document}